\documentclass{article}

\usepackage{arxiv}

\usepackage[authoryear,round]{natbib}

\usepackage[utf8]{inputenc}
\usepackage[T1]{fontenc}
\usepackage{amsmath}
\usepackage{amssymb}
\usepackage{amsthm}
\usepackage[ruled,vlined]{algorithm2e}
\usepackage{booktabs}
\usepackage{multirow}
\usepackage{tabularx}
\usepackage{array}
\usepackage{enumitem}
\usepackage{xspace}
\usepackage{xcolor}
\usepackage{graphicx}
\usepackage{float}
\usepackage{listings}
\usepackage{tikz}
\usetikzlibrary{arrows.meta,positioning,fit,backgrounds,calc,shadows}
\usepackage{fontawesome5}
\pdfmapfile{+fontawesome5.map}
\PassOptionsToPackage{hyphens}{url}
\usepackage[hidelinks]{hyperref}
\usepackage{url}

\definecolor{cData}{HTML}{2563EB}   % blue   -- per-dataset stage
\definecolor{cExpl}{HTML}{0F766E}   % teal   -- explore stage
\definecolor{cLoop}{HTML}{C2410C}   % orange -- main-loop stage
\definecolor{cAns}{HTML}{15803D}    % green  -- final answer
\definecolor{cProbe}{HTML}{B45309}  % amber  -- discover/probe ops
\definecolor{cRule}{HTML}{7C3AED}   % violet -- rule engine

\theoremstyle{definition}

\newcommand{\system}{RADAR\xspace}
\newcommand{\cnum}[1]{\tikz[baseline=-0.55ex]{\node[circle,fill=black!75,text=white,inner sep=0.9pt,font=\bfseries\scriptsize]{#1};}}

\lstdefinestyle{raat}{
  basicstyle=\ttfamily\footnotesize,
  breaklines=true,
  columns=fullflexible,
  keepspaces=true,
  showstringspaces=false,
  frame=single,
  framesep=4pt,
  xleftmargin=6pt,
  aboveskip=6pt,belowskip=6pt,
}
\newcolumntype{Y}{>{\raggedright\arraybackslash}X}

\title{Fail Loudly: An Auditable Runtime for Agentic Data Analysis}
\author{
  {\normalfont\fontsize{13}{16}\selectfont \href{https://orcid.org/0009-0000-0140-6350}{Hanxu Yan}}\\[2pt]
  {\normalfont\fontsize{11}{13}\selectfont Independent Researcher}\\[1pt]
  {\normalfont\fontsize{11}{13}\selectfont\href{mailto:hanxuyan888@gmail.com}{hanxuyan888@gmail.com}}
  \And
  {\normalfont\fontsize{13}{16}\selectfont \href{https://orcid.org/0009-0007-3016-8700}{Langxuan Deng}}\\[2pt]
  {\normalfont\fontsize{11}{13}\selectfont Independent Researcher}\\[1pt]
  {\normalfont\fontsize{11}{13}\selectfont\href{mailto:langxvnadeng@gmail.com}{langxvnadeng@gmail.com}}
  \AND
  {\normalfont\fontsize{13}{16}\selectfont \href{https://orcid.org/0009-0003-0221-6761}{Zhengle Wang}}\\[2pt]
  {\normalfont\fontsize{11}{13}\selectfont Purdue University}\\[1pt]
  {\normalfont\fontsize{11}{13}\selectfont\href{mailto:wang7690@purdue.edu}{wang7690@purdue.edu}}
  \And
  {\normalfont\fontsize{13}{16}\selectfont \href{https://orcid.org/0009-0005-1971-3398}{Yibo Wang}}\\[2pt]
  {\normalfont\fontsize{11}{13}\selectfont Purdue University}\\[1pt]
  {\normalfont\fontsize{11}{13}\selectfont\href{mailto:wang7342@purdue.edu}{wang7342@purdue.edu}}
  \And
  {\normalfont\fontsize{13}{16}\selectfont \href{https://orcid.org/0000-0003-1481-2678}{Chunwei Liu}\textsuperscript{\(\dagger\)}}\\[2pt]
  {\normalfont\fontsize{11}{13}\selectfont Purdue University}\\[1pt]
  {\normalfont\fontsize{11}{13}\selectfont\href{mailto:chunwei@purdue.edu}{chunwei@purdue.edu}}
}
\date{}
\renewcommand{\shorttitle}{Fail Loudly: An Auditable Runtime for Agentic Data Analysis}
\renewcommand{\headeright}{}
\hypersetup{
  pdftitle={Fail Loudly: An Auditable Runtime for Agentic Data Analysis},
  pdfauthor={Hanxu Yan, Langxuan Deng, Zhengle Wang, Yibo Wang, Chunwei Liu},
  pdfkeywords={agentic data analysis, runtime validation, silent errors, RADAR}
}

\begin{document}
\raggedbottom
\maketitle

\begin{abstract}
Large language models (LLMs) have enabled data-science agents to automate multi-step analyses over heterogeneous files. However, incorrect choices regarding data sources, scope, or statistical definitions often lead to silent errors: computations execute successfully but produce plausible yet incorrect outputs that fail to answer the intended question. To mitigate this, we present \system, an auditable runtime that makes an agent's analytical choices inspectable and supports their revision through execution feedback. \system operates through three core mechanisms. First, an evidence-preserving exploration module retrieves task-relevant content while retaining source locations and observation coverage.
Next, the runtime uses typed operators to record the agent's declared inputs, operation arguments, and resulting observations.
Finally, runtime validation checks proposed operations against these observations.
When a conflict is detected, the runtime rejects the operation or provides diagnostic feedback, allowing the agent to revise its choices before errors propagate.
This design enables agents to fail loudly while leaving semantic interpretation to the LLM. On KramaBench, \system achieves overall scores of 0.723 with full source retrieval and 0.747 with gold sources supplied, corresponding to relative gains of 35.9\% and 28.8\% over the strongest baselines. Beyond KramaBench, \system achieves relative performance gains of 14.0\% on DA-Code and 59.3\% on DABStep, demonstrating its applicability across diverse agentic data-analysis workflows.

\end{abstract}

\begin{figure}[t]
  \centering
  \includegraphics[width=\linewidth]{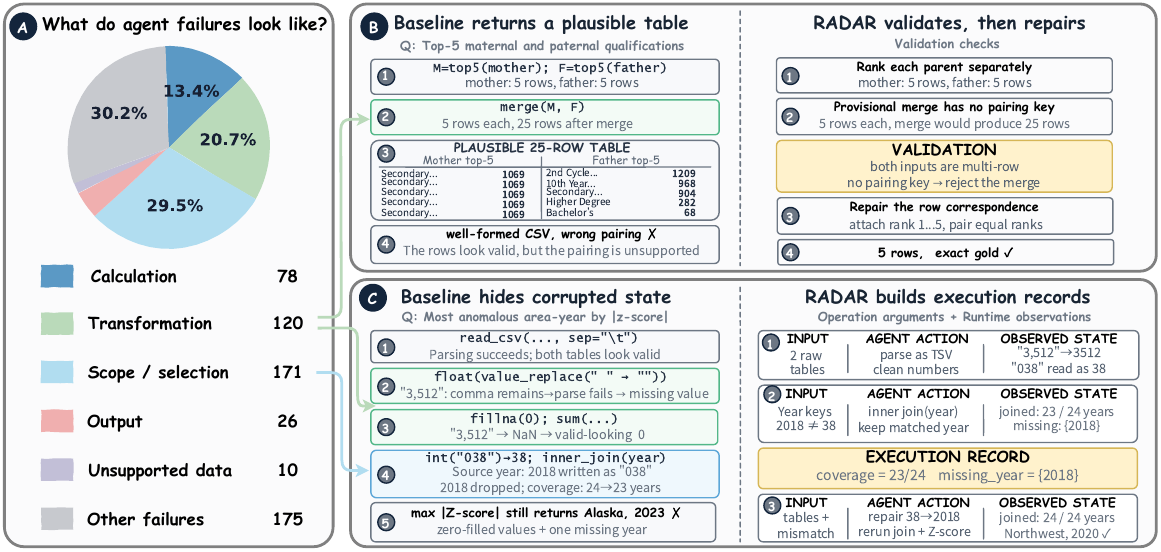}
  \caption{Silent errors and runtime interventions by \system. (A) Among 580 zero-scored DS-STAR runs across three benchmarks,
  405 (69.8\%) return normal-looking but incorrect outputs; we categorize their primary error types
  alongside other failures. (B) A baseline combines two Top-5 rankings without a pairing key,
  producing a well-formed 25-row table with incorrect row correspondence. On the same task, \system
  rejects an unkeyed merge before execution, prompting the agent to pair entries by rank. (C) A baseline
  silently loses one year after a successful inner join, while \system records the agent's
  inspection output that reveals the missing year and guides repair.}
  \label{fig:motivation}
  \vspace{-6pt}
\end{figure}

\section{Introduction}
\label{sec:intro}

LLM-based data-science agents can automate analyses over heterogeneous files, but successful code
execution does not establish that an analysis answers the intended question. These agents interpret
natural-language tasks, locate and combine data, and execute code to produce answers and
artifacts~\citep{react,autogen,datacopilot,dsagent,datainterp,dsstar}. Translating a natural-language question into
code requires commitments about the source, scope, field semantics, and statistical definition~\citep{automatingds}.
Incorrect choices can produce silent errors: plausible but incorrect outputs delivered without
an explicit indication of failure~\citep{flowcheck}.
Unlike exceptions or missing files, these errors need not interrupt execution, allowing later
computations to proceed on an incorrect basis~\citep{stoisser2026ambigdsbenchmarktaskframingambiguity}.

Our audit shows that silent errors persist even in an agent equipped with iterative verification.
Among 580 zero-scored DS-STAR~\citep{dsstar} runs across KramaBench, DA-Code, and
DABStep~\citep{kramabench,dacode,dabstep}, 405 ($69.8\%$) delivered normal-looking but incorrect results
without reporting failure (Figure~\ref{fig:motivation}A). For example, a baseline asked to place two
Top-5 rankings side by side merges them without a pairing condition (Figure~\ref{fig:motivation}B).
The operation succeeds and produces a well-formed 25-row table with incorrect row correspondence.
Such failures call for feedback that identifies suspect analytical choices despite successful execution.

Existing approaches largely rely on models to interpret source information and assess an analysis,
leaving room for explicit runtime checks on individual operations. File descriptions and specialized
file agents help navigate heterogeneous collections within a bounded context~\citep{dsstar,blackboardds}.
Recent work iteratively verifies plans using code and execution results or probes intermediate
states to assess silent and grounding errors~\citep{dsstar,dataprm}. Checking an individual choice requires
linking it to the relevant source observations and their coverage. When summaries omit critical
constraints or these links are lost between analysis steps, unsupported interpretations can remain
unchallenged as later computations depend on them~\citep{laban2026llms}. The challenge is to preserve this context across
exploration and execution, and use it to determine when an operation warrants intervention.

Our key insight is that selected analytical choices become checkable when linked to source evidence
and runtime observations with explicit scope and coverage. For example, a complete enumeration of a
column's values can establish that a proposed filter value is absent, whereas its omission from a
partial preview cannot. In the ranking example (Figure~\ref{fig:motivation}B), the missing pairing condition and recorded input
sizes provide grounds to reject an implicit join and request an explicit correspondence. Such checks
identify local conflicts, while the agent remains responsible for interpreting the task and choosing
an appropriate revision.

To operationalize this, we present \system, an auditable runtime that connects evidence-preserving
exploration, typed execution, and runtime validation. Evidence-preserving exploration combines
retrieval and profiling with LLM semantic selection, returning selected content
with source locations, observation scope and coverage.
Typed operators expose operation arguments
and declared dependencies, and their execution records link each invocation to observations of its inputs
and results. Runtime validation uses these records to check operation arguments, execution results, and output requirements
against the available evidence. On conflict, it responds through rejection, targeted diagnostic feedback, or
bounded internal repair. This design separates semantic interpretation from runtime validation,
enabling agents to fail loudly and revise their analysis.

Across KramaBench, DA-Code, and DABStep, \system improves task performance while reducing both
observed silent errors and non-delivery. Across 1,054 tasks, silent errors decrease from 405 for
DS-STAR to 289 for \system, a reduction of $28.6\%$, while non-deliveries decrease from 84 to 24.
On KramaBench, it achieves Overall scores of $0.723$ under Full Input and $0.747$ under Oracle Input,
compared with $0.532$ and $0.580$ for the strongest baseline in each regime. The advantage with gold
sources supplied indicates that the gains extend beyond source retrieval. On DA-Code and DABStep,
it improves over DS-STAR, the strongest baseline on both benchmarks, by $6.5$ and $19.5$ points,
respectively. Ablations show that both evidence-preserving exploration and runtime validation
contribute to higher task scores and fewer silent errors.

\section{Evidence-Preserving Exploration}
\label{sec:exploration}

\begin{figure}[t]
  \centering
  \includegraphics[width=\linewidth]{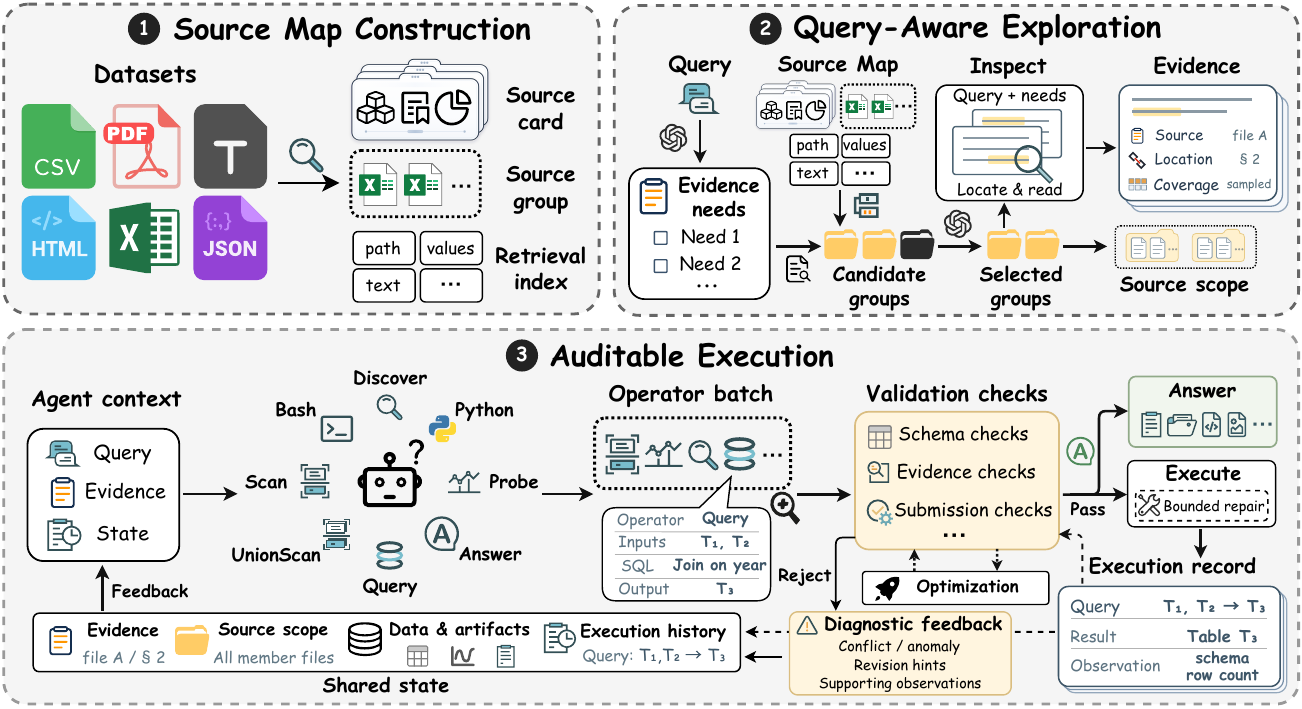}
  \caption{Overview of \system. Source Map construction supports query-aware source selection and inspection. The agent submits typed operator batches to the runtime, which applies
  validation and optimization, records runtime observations in shared state, and returns feedback
  for revision.}
  \label{fig:overview}
  \vspace{-10pt}
\end{figure}

\system retrieves evidence for analysis while preserving its source locations and observation
coverage. The input is a natural-language query $q$ over a data lake of heterogeneous files.
Source Map Construction in Figure~\ref{fig:overview}\,\cnum{1} builds a query-independent
representation of these files for retrieval.
Query-Aware Exploration in Figure~\ref{fig:overview}\,\cnum{2} then selects relevant sources
and inspects their contents using $q$.
The outputs are the retrieved evidence $E_q$ and the selected source scope $\mathcal{X}_q$,
which identifies the files retained for execution. Both initialize the analysis in
Section~\ref{sec:auditable-execution}.

\subsection{Source Map Construction}
\label{sec:source-map}

Source Map construction records what each file contains and where its contents can be found,
so these observations can be reused across queries.
Because a file may contain several addressable parts, such as workbook sheets, \system
profiles these parts separately as source units.
Each unit retains its file path and position, such as a sheet name, table index, or JSON pointer.
A format-specific profiler records each unit's observed structure and representative content,
including schemas and value characteristics, document sections, or array shapes and types.
The profiler stores these descriptions in a per-file record called a Source Card.
Each card also records observation coverage, distinguishing complete inspection from observations
limited by the profiling budget.

Related files or units are grouped so that retrieval and selection can consider them together.
\system forms these source groups from units in the same workbook, files with shared path
patterns, and compatible tables in the same directory. For example, files partitioned by year can be retrieved as a
group so that source selection can consider multiple years together.
These groups form the set $\mathcal{G}$, with each remaining source assigned to its own group.
To make the recorded information searchable, \system builds a retrieval index over paths and
names, observed columns and values, and descriptive content.

\subsection{Query-Aware Exploration}
\label{sec:source-selection}
\label{sec:evidence-handoff}

Query-aware exploration uses the information needed to answer $q$ to guide source retrieval,
selection, and inspection. An LLM first decomposes $q$ into evidence needs, each describing
one piece of source information required by the question.
Each retrieval view $v$ contains either $q$ alone or $q$ paired with one evidence need.
For each view, BM25~\citep{bm25} ranks source groups separately for each indexed field
$f\in\mathcal{F}$.
A group at position $r$ in a field ranking receives a contribution of $1/(\kappa+r)$,
where $\kappa>0$ is an offset. We sum these contributions across fields to obtain
\[
    s_v(g)=\sum_{f\in\mathcal{F}}
    \frac{1}{\kappa+\operatorname{rank}_{v,f}(g)},
\]
where $\operatorname{rank}_{v,f}(g)$ is group $g$'s position in field $f$'s ranking for view $v$.
Combining ranks prioritizes groups that rank highly in multiple fields without requiring BM25
scores from different fields to be directly comparable.

Candidate construction retains groups retrieved for different evidence needs.
\system interleaves the rankings from the retrieval views to form the candidate set
$\mathcal{K}_q\subseteq\mathcal{G}$, keeping the first occurrence of each group.
Retrieval over individual source units also highlights relevant members of each candidate group.
For each candidate, the LLM selector receives member paths, observed fields and values,
content previews from Source Cards, and matched document sections.
Using $q$ and its evidence needs, the selector chooses groups likely to contain the required
evidence, forming $\mathcal{G}_q\subseteq\mathcal{K}_q$.

\system inspects the selected groups using $q$ and its evidence needs to retrieve relevant
records and passages.
For tables, \system combines structural information from Source Cards with original rows
selected by lexical and semantic matching.
For document text, it ranks sections and passages by relevance and retains adjacent context
for lists and tables. The resulting evidence $E_q$ identifies file paths and positions within
each file, allowing the agent to return to the underlying content.
Accompanying reports identify which sources were inspected, which portions are presented,
and content omitted because of inspection limits.

The handoff to execution preserves all files in the selected groups, including files that
contributed no displayed evidence. The selected source scope is
\[
    \mathcal{X}_q=\bigcup_{g\in\mathcal{G}_q}\operatorname{Files}(g),
\]
where $\operatorname{Files}(g)$ denotes the files belonging to group $g$.
The Evidence and Source scope outputs in Figure~\ref{fig:overview}\,\cnum{2} therefore serve
different roles, providing observed content and retaining files for further analysis.
The agent receives $(E_q,\mathcal{X}_q)$ as its initial context, and the runtime retains
both in the shared state shown in Figure~\ref{fig:overview}\,\cnum{3}.
Subsequent operations can load the selected sources and inspect additional content,
adding observations beyond the initial evidence.

\section{Auditable Execution}
\label{sec:auditable-execution}

Auditable execution uses the evidence and source scope from exploration to check operations, record invocations and runtime observations, and support revision (Figure~\ref{fig:overview}\,\cnum{3}). Using the initial context, the agent submits operator batches whose structured arguments and declared inputs expose its analytical choices to the runtime. Validation checks use the available evidence to determine whether to reject an invocation or allow it to proceed. Execution observations and failures can trigger diagnostic feedback or bounded internal repair. Execution records and diagnostic feedback update shared state and inform the agent's next batch.

\subsection{Typed Operators and Execution Records}
\label{sec:operator-interfaces}
\label{sec:execution-records}

Typed operators expose the arguments and input references needed to connect each operation to relevant observations. Each type specifies structured arguments, an execution procedure, and the observations it returns. Table~\ref{tab:operators} summarizes the eight operator types, which cover source discovery, data access, inspection, computation, and answer submission. For example, Query executes SQL through DuckDB~\citep{duckdb}, with \texttt{inputs} naming its input tables and \texttt{output} naming the result. The Query illustrated in Figure~\ref{fig:overview} joins $T_1$ and $T_2$ on year and names its result $T_3$. The declared inputs let the runtime check the SQL against observations of the referenced tables, while the output name allows subsequent operations to reference the result.

\begin{table}[t]
\centering
\small
\setlength{\tabcolsep}{2pt}
\caption{Typed operators, their main arguments, and runtime observations.}
\label{tab:operators}
\begin{tabular*}{\linewidth}{@{\extracolsep{\fill}}lll@{}}
\toprule
Operator type & Main arguments & Function and runtime observations \\
\midrule
Discover & \texttt{query}, \texttt{excluded} & Retrieve sources by semantic query or path pattern; return paths. \\
Bash & \texttt{command} & Inspect files with shell commands; record output and exit status. \\
Scan & \texttt{file} & Load a table; record provenance, schema, row count, and preview. \\
UnionScan & \texttt{files}, \texttt{source\_column} & Union by column name; record source labels and table metadata. \\
Probe & \texttt{table}, \texttt{columns} & Inspect columns; report value frequencies, ranges, and null rates. \\
Query & \texttt{sql}, \texttt{inputs}, \texttt{output} & Run SQL; record result schema, row count, and visible records. \\
Python & \texttt{code}, \texttt{inputs}, \texttt{output} & Run code; record values, program output, and artifact metadata. \\
Answer & \texttt{result\_type}, \texttt{result} & Submit an explicit result; check its type and artifact requirements. \\
\bottomrule
\end{tabular*}
\end{table}

After execution, the runtime creates an execution record containing the invocation's arguments, declared inputs, and resulting observations. Shared state retains these records together with exploration evidence, the selected source scope, loaded data, and artifact metadata. When an invocation names $T_3$ as an input, the runtime can retrieve its recorded schema and row count, along with any Probe observations for its columns. The agent may submit several invocations in one response, forming a batch that the runtime processes in order. Records and diagnostic feedback update shared state after each invocation, so an earlier Probe can supply evidence for a later Query in the same batch.

\subsection{Validation Checks}
\label{sec:runtime-validation}

The Validation checks stage in Figure~\ref{fig:overview} compares choices expressed in operation arguments with observations of the referenced inputs. Algorithm~\ref{alg:validation-checks} shows how the runtime checks a Query invocation. Each check requires a supported SQL structure and the observations specified by its rule, and is skipped when either requirement is unmet. If none rejects the invocation, it follows the Pass branch and proceeds to execution, while the agent remains responsible for deciding whether the analysis answers the question.

\begin{algorithm}[!b]
\small
\SetAlFnt{\small}
\DontPrintSemicolon
\SetKw{Continue}{continue}
\SetKwFor{ForEach}{for each}{do}{end}
\caption{Pre-execution checks for a Query invocation}
\label{alg:validation-checks}
\KwIn{Query invocation $o$ and current shared state $S$}
\ForEach{Query check $r$}{
  \lIf{$r$ does not support the SQL structure of $o$}{\Continue}
  $e \leftarrow$ retrieve the observations required by $r$ for the inputs of $o$ from $S$\;
  \lIf{$e$ does not meet the evidence requirements of $r$}{\Continue}
  \If{the rejection condition of $r$ holds for $(o,e)$}{
    \Return{Reject with the conflict, supporting observations, and revision hints}\;
  }
}
\Return{Proceed to execution}\;
\end{algorithm}

Observation coverage determines whether an unobserved filter value provides grounds for rejection. An exploration preview identifies inspected content, while the literal coverage check requires a complete value enumeration from Probe. For a supported equality predicate $T.c=v$, let $L_{T,c}$ be Probe's returned list of non-null values for table $T$ and text column $c$. With $d_{T,c}$ denoting the distinct non-null count and $\tau$ the enumeration bound, the required completeness condition is
\[
\operatorname{Complete}(T,c)
\iff 0<d_{T,c}\leq\tau
\;\land\;
|L_{T,c}|=|\operatorname{set}(L_{T,c})|=d_{T,c}.
\]
The check rejects a string literal $v$ absent from $\operatorname{set}(L_{T,c})$ only when this condition holds. For example, if the complete list contains only \texttt{USA} and \texttt{Canada}, the check rejects \texttt{country = 'United States'} and returns the observed values. The same omission from a partial preview leaves the check without sufficient evidence to reject.

Schema, evidence, and submission checks cover different sources of conflict. Within its supported cases, the field binding check rejects a column name absent from all known input schemas and query-defined aliases. For an implicit join between multirow inputs, the runtime requests an explicit pairing condition or an explicitly intended cross product. In the ranking example (Figure~\ref{fig:motivation}B), combining two five-row inputs without a pairing condition triggers this rule before the operation produces 25 combinations. Answer checks the declared result type and applicable artifact requirements before accepting the result and ending the analysis.

\subsection{Diagnostic Feedback and Bounded Repair}
\label{sec:validation-feedback}

Diagnostic feedback supports revision after a rejected invocation or a suspicious execution result. The Reject branch in Figure~\ref{fig:overview} returns the conflict, supporting observations, and revision hints. For invocations that proceed to execution, the resulting observations may also warrant diagnostic feedback. For example, when a filtered or joined Query succeeds but returns no rows, the runtime inspects the referenced values and join keys and returns diagnostics alongside the result. Since an empty result can be valid, the agent decides whether to accept it or revise the operation. Advisory rules use shared state to guide the next batch through the Feedback path without rejecting an invocation. Revised invocations undergo the same checks.

The Execute stage may attempt bounded internal repair for SQL execution failures and failed or suspicious data loads, using an LLM with the relevant error and source observations. For SQL repair, the prompt requires preserving analytical intent. The runtime rejects a candidate if its underlying data sources differ from those of the submitted SQL or either source set cannot be extracted. Execution records retain both SQL versions; failed repair returns diagnostics to the main agent.

\section{Experiments}
\label{sec:eval}

\subsection{Experimental Setup}

\noindent\textbf{Benchmarks.} We evaluate on KramaBench~\citep{kramabench}, DA-Code~\citep{dacode},
and DABStep~\citep{dabstep}. KramaBench contains $104$ tasks over heterogeneous data lakes spanning
six domains. We consider Full Input, which requires source retrieval, and Oracle Input,
which supplies the gold sources. DA-Code contains $500$ tasks covering Data Wrangling, Machine
Learning (ML), and exploratory data analysis (EDA). DABStep contains $450$ tasks over structured
and unstructured financial data.

\noindent\textbf{Baselines.} We compare six agent-based solutions: DS-STAR's file analysis and iterative
verification~\citep{dsstar}; DA-Agent's reasoning--action--observation loop~\citep{dacode};
Data Interpreter's task graphs and progressive verification~\citep{datainterp};
AutoGen's multi-agent coordination~\citep{autogen}; smolagents Deep Research~\citep{smolagents};
and DS-Guru, KramaBench's provided baseline~\citep{kramabench}. The KramaBench human expert
study is included for reference.

\noindent\textbf{Implementation.} All systems receive the same raw files within each input setting and use their
own exploration pipelines. The main experiments use DeepSeek-V4-Flash (Preview) for all
systems with temperature $0$. All LLM components in \system use this model.
We allow all systems up to $25$ main-agent decision rounds per task.
We report total query-time API cost
over each evaluated task suite and mean per-task latency, which includes offline preprocessing time amortized across queries.

\subsection{Main Results}
\label{sec:e2e}
\label{sec:dacode}
\label{sec:dabstep}

\system achieves the highest Overall scores across all three benchmarks with DeepSeek-V4-Flash (Tables~\ref{tab:main}, \ref{tab:dacode}, and \ref{tab:dabstep}). On KramaBench, it leads all six domains under both Full Input and Oracle Input. With gold sources supplied, its Overall score reaches $0.747$, compared with $0.580$ for DS-STAR, the strongest baseline in this setting. This result indicates that the gains extend beyond source retrieval. On DA-Code, \system improves across Data Wrangling, ML, and EDA, with its largest margin in Data Wrangling. This pattern is consistent with its emphasis on inspecting source data and checking transformations during execution. On DABStep, the Overall improvement is concentrated in Hard tasks, where \system achieves $47.1\%$ versus $22.8\%$ for DS-STAR, an increase of $24.3$ percentage points. These tasks account for $84\%$ of the benchmark, so their gains outweigh a $5.5$-point decrease on Easy tasks.

\begin{table*}[t]
\centering
\scriptsize
\setlength{\tabcolsep}{3pt}
\caption{End-to-end scores and resource use on KramaBench across backbone models.
Overall is the task-count-weighted mean. The Human row is from the KramaBench expert study.
Bold marks the best automated system within each input setting and backbone.
GPT-5.6 Luna uses high reasoning effort; GPT-4.1-mini uses the 2025-04-14 snapshot.} 
\label{tab:main}
\resizebox{\linewidth}{!}{%
\begin{tabular}{@{}lcccccccccc@{}}
\toprule
& & \multicolumn{7}{c}{Domains} & \multicolumn{2}{c}{Resource use} \\
\cmidrule(lr){3-9}\cmidrule(l){10-11}
System & Models & Archaeology & Astronomy & Biomedical & Environment & Legal & Wildfire & Overall & Cost (\$) & Latency (s) \\
\midrule
\multicolumn{11}{c}{Full Input (source retrieval from the complete data lake)} \\
\midrule
Human    &  & 0.583 & 0.700 & 1.000 & 0.769 & 0.867 & 0.662 & 0.768 & --   & --  \\
\midrule
DS-Guru          & \multirow{7}{*}{\shortstack{DeepSeek\\V4-Flash}} & 0.000 & 0.083 & 0.000 & 0.150 & 0.200 & 0.305 & 0.158 & 1.27 & 49  \\
smolagents DR    & & 0.417 & 0.250 & 0.556 & 0.533 & 0.533 & 0.746 & 0.532 & 6.38 & 636 \\
AutoGen          & & 0.417 & 0.250 & 0.111 & 0.333 & 0.500 & 0.380 & 0.372 & 3.16 & 183 \\
Data Interpreter & & 0.250 & 0.167 & 0.000 & 0.300 & 0.200 & 0.681 & 0.301 & 0.80 & 170 \\
DA-Agent         & & 0.417 & 0.208 & 0.111 & 0.469 & 0.567 & 0.651 & 0.467 & 1.32 & 105 \\
DS-STAR          & & 0.333 & 0.167 & 0.500 & 0.250 & 0.667 & 0.708 & 0.484 & 3.47 & 307 \\
\textbf{\system}  & & \textbf{0.500} & \textbf{0.625} & \textbf{0.667}
                 & \textbf{0.733} & \textbf{0.800} & \textbf{0.809}
                 & \textbf{0.723} & 1.16 & 89 \\
\midrule
DS-Guru          & \multirow{7}{*}{\shortstack{GPT-5.6\\Luna}} & 0.333 & 0.083 & 0.111 & 0.150 & 0.250 & 0.309 & 0.221 & 2.08 & 301 \\
smolagents DR    & & 0.333 & 0.500 & 0.444 & 0.700 & 0.396 & 0.599 & 0.504 & 7.44 & 476 \\
AutoGen          & & \textbf{0.417} & 0.167 & 0.444 & 0.460 & 0.433 & 0.475 & 0.415 & 4.22 & 49 \\
Data Interpreter & & \textbf{0.417} & 0.250 & 0.444 & 0.200 & 0.233 & 0.309 & 0.283 & 1.69 & 70 \\
DA-Agent         & & \textbf{0.417} & 0.417 & 0.556 & 0.683 & 0.667 & 0.703 & 0.610 & 1.32 & 80 \\
DS-STAR          & & 0.333 & 0.250 & 0.444 & 0.500 & 0.733 & 0.578 & 0.530 & 7.87 & 188 \\
\textbf{\system} & & \textbf{0.417} & \textbf{0.708} & \textbf{0.667}
                & \textbf{0.783} & \textbf{0.853} & \textbf{0.761}
                & \textbf{0.738} & 1.24 & 75 \\
\midrule

DS-Guru          & \multirow{7}{*}{\shortstack{GPT-4.1\\mini}} & 0.083& 0.000& 0.000&0.000 &0.067 & 0.179& 0.065& 1.29& 24\\
smolagents DR    & & 0.167& 0.125& 0.222& 0.350&0.250 & 0.319& 0.257& 6.22& 77\\
AutoGen          & & 0.250& 0.271& 0.167& 0.307& 0.331& 0.388& 0.307& 4.76& 45\\
Data Interpreter & & 0.167&0.167 & 0.000& 0.350&0.220 &0.429 & 0.256& 4.28& 249\\
DA-Agent         & & 0.250 & 0.208 & 0.111 & 0.380 & 0.430 & 0.386 & 0.338 & 1.91 & 56 \\
DS-STAR          & & 0.167 & 0.250 & 0.222 & 0.413 & 0.420 & \textbf{0.453} & 0.359 & 6.44 & 76 \\
\textbf{\system} & & \textbf{0.417} & \textbf{0.333} & \textbf{0.278}
                & \textbf{0.493} & \textbf{0.600} & 0.406
                & \textbf{0.460} & 2.29 & 47 \\
\midrule
\multicolumn{11}{c}{Oracle Input (gold sources supplied)} \\
\midrule
DS-Guru          & \multirow{7}{*}{\shortstack{DeepSeek\\V4-Flash}} & 0.167 & 0.083 & 0.111 & 0.150 & 0.200 & 0.286 & 0.183 & 0.72 & 43  \\
smolagents DR    & & 0.417 & 0.333 & 0.667 & 0.402 & 0.600 & 0.746 & 0.545 & 4.27 & 346 \\
AutoGen          & & 0.417 & 0.417 & 0.556 & 0.452 & 0.633 & 0.535 & 0.522 & 1.55 & 62  \\
Data Interpreter & & 0.333 & 0.250 & 0.222 & 0.319 & 0.333 & 0.681 & 0.382 & 0.52 & 101 \\
DA-Agent         & & 0.333 & 0.375 & 0.333 & 0.652 & 0.600 & 0.503 & 0.511 & 0.94 & 95  \\
DS-STAR          & & 0.333 & 0.417 & 0.556 & 0.508 & 0.700 & 0.719 & 0.580 & 0.93 & 245 \\
\textbf{\system}  & & \textbf{0.583} & \textbf{0.667} & \textbf{0.778}
                 & \textbf{0.783} & \textbf{0.800} & \textbf{0.764}
                 & \textbf{0.747} & 0.91 & 67 \\
\bottomrule
\end{tabular}%
}
\end{table*}

\begin{table*}[t]
\centering
\small
\setlength{\tabcolsep}{3.5pt}
\caption{DA-Code mean scores (\%) on 500 tasks. EDA aggregates 79 Data Insights,
73 Data Manipulation, 70 Statistical Analysis, and 78 Visualization tasks, weighted by task
count. Completion represents the proportion of tasks for which the model produces results.}
\label{tab:dacode}
\begin{tabular}{@{}lcccccccc@{}}
\toprule
System & Data Wrangling & ML & EDA & Easy & Medium & Hard & Overall & Completion \\
\midrule
DS-Guru        & 16.5 & 34.6 & 16.5 & 27.5 & 18.5 & 17.3 & 20.1 & 70.8 \\
smolagents DR  & 29.2 & 44.8 & 39.6 & 53.8 & 36.6 & 28.7 & 38.5 & 91.6 \\
AutoGen        & 31.4 & 41.8 & 35.2 & 46.8 & 33.7 & 30.7 & 35.8 & 86.6 \\
Data Interpreter & 21.0 & 24.5 & 32.6 & 36.8 & 26.9 & 25.3 & 28.6 & 84.8 \\
DA-Agent       & 33.9 & 51.5 & 44.3 & 55.4 & 42.1 & 36.3 & 43.6 & 90.0 \\
DS-STAR        & 33.6 & 58.8 & 46.6 & 61.4 & 42.6 & 42.2 & 46.4 & 96.8 \\
\textbf{\system} & \textbf{42.4} & \textbf{62.6} & \textbf{53.2}
               & \textbf{69.0} & \textbf{50.4} & \textbf{43.9} & \textbf{52.9} & \textbf{98.4} \\
\bottomrule
\end{tabular}
\end{table*}

\subsection{Effect on Silent Errors}
\label{sec:silent-errors}

To assess silent errors beyond benchmark scores, we audit zero-scored runs across the three benchmarks and group silent errors into five categories:
wrong scope or selection, wrong transformation, wrong calculation, wrong output, and fabricated
or unsupported data. Each silent error is assigned one primary category; remaining failed runs
are classified as other failures.

\begin{figure}[!hb]
  \centering
  \includegraphics[width=\linewidth]{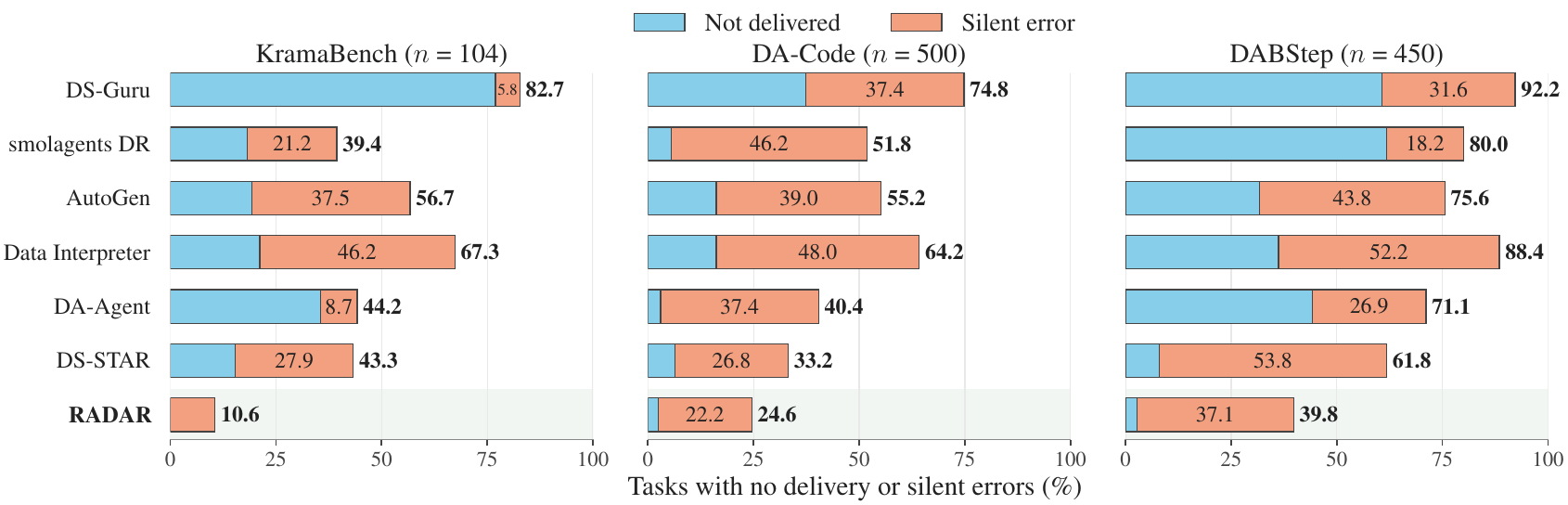}
  \caption{Non-delivery and silent errors across systems on each benchmark.}
  \label{fig:silent-error-comparison}
\end{figure}

Figure~\ref{fig:silent-error-comparison} shows that \system has the lowest combined rate of
non-delivery and silent errors on each benchmark. Compared with DS-STAR, it reduces both
outcomes across all three benchmarks. Across 1,054 tasks, silent errors decrease from 405 to
289, a reduction of $28.6\%$, while non-deliveries decrease from 84 to 24. The combined rate is
$29.7\%$, compared with $46.4\%$--$83.0\%$ for the baselines.

\begin{table}[t]
  \begin{minipage}[t]{0.52\linewidth}
    \vspace{0pt}
    \centering
    \small
    \setlength{\tabcolsep}{10pt}
    \caption{DABStep accuracy (\%). Easy, Hard, and Overall cover $72$, $378$, and $450$ tasks,
    respectively; \textbf{bold} marks the best result per column.}
    \label{tab:dabstep}
    \begin{tabular}{@{}lccc@{}}
    \toprule
    System & Easy & Hard & Overall \\
    \midrule
    DS-Guru & 45.8 & 0.5 & 7.8  \\
    smolagents DR & 73.6 & 9.8 & 20.0 \\
    AutoGen & 69.4 & 15.9 & 24.4 \\
    Data Interpreter & 66.7 & 1.1 & 11.6 \\
    DA-Agent       & 75.0 & 20.1 & 28.9 \\
    DS-STAR        & \textbf{86.1} & 22.8 & 32.9 \\
    \textbf{\system} & 80.6 & \textbf{47.1} & \textbf{52.4} \\
    \bottomrule
    \end{tabular}
  \end{minipage}\hfill
  \begin{minipage}[t]{0.46\linewidth}
    \vspace{0pt}
    \centering
    \includegraphics[width=\linewidth]{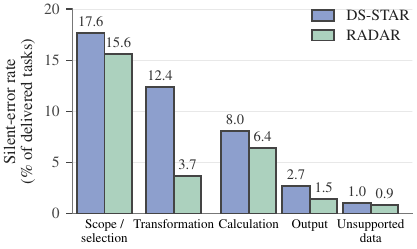}
    \makeatletter\def\@captype{figure}\makeatother
    \caption{Silent-error rates by category among delivered tasks for DS-STAR and \system.}
    \label{fig:silent-error-rates-delivered}
  \end{minipage}
\end{table}

Among substantive deliveries, the silent-error rate decreases from $41.8\%$ for DS-STAR
 to $28.1\%$ for \system (Figure~\ref{fig:silent-error-rates-delivered}).
Transformation errors show the largest absolute decrease, from $12.4\%$ to $3.7\%$, consistent with
the runtime's checks on input relationships and execution results. Output errors decrease by
$45.7\%$, aligning with output-contract validation. Scope or selection remains the most
frequent category, decreasing from $17.6\%$ to $15.6\%$.

\subsection{Component Analysis}
\label{sec:components}

We ablate evidence-preserving exploration and runtime validation across all three benchmarks.
The exploration ablation uses only file metadata and provides no initial
table or document evidence. Without runtime validation, checks, feedback, and internal repair
are disabled.

\begin{table}[t]
  \begin{minipage}[t]{0.50\linewidth}
    \vspace{0pt}
    \centering
    \footnotesize
    \setlength{\tabcolsep}{3pt}
    \renewcommand{\arraystretch}{1.15}
    \caption{Ablation results across the three benchmarks. Parentheses show decreases in percentage
    points from Full.}
    \label{tab:component-ablation}
    \begin{tabular}{@{}lccc@{}}
      \toprule
      Setting & KramaBench & DA-Code & DABStep \\
      \midrule
      Full & \textbf{72.3} & \textbf{52.9} & \textbf{52.4} \\
      \addlinespace[3pt]
      $\vcenter{\hbox{\shortstack[l]{w/o evidence-\\preserving\\exploration}}}$
        & $\vcenter{\hbox{\shortstack{$61.6$\\$(-10.7)$}}}$
        & $\vcenter{\hbox{\shortstack{$48.9$\\$(-4.0)$}}}$
        & $\vcenter{\hbox{\shortstack{$44.2$\\$(-8.2)$}}}$ \\
      \addlinespace[3pt]
      \shortstack[l]{w/o runtime\\validation}
        & \shortstack{$64.6$\\$(-7.7)$}
        & \shortstack{$50.6$\\$(-2.3)$}
        & \shortstack{$42.2$\\$(-10.2)$} \\
      \bottomrule
    \end{tabular}
  \end{minipage}\hfill
  \begin{minipage}[t]{0.48\linewidth}
    \vspace{0pt}
    \centering
    \includegraphics{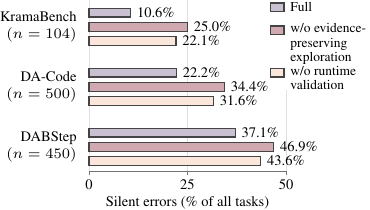}\hspace{2.5pt}
    \makeatletter\def\@captype{figure}\makeatother
    \caption{Silent-error rates under component ablations.}
    \label{fig:component-silent-errors}
  \end{minipage}
\end{table}

Table~\ref{tab:component-ablation} shows that both ablations reduce performance across
all three benchmarks. Removing evidence-preserving exploration has the largest effect
 on KramaBench, consistent with its need to locate and interpret sources
across heterogeneous data lakes. DABStep shows the largest decrease from removing
runtime validation, consistent with the need to check intermediate decisions across multi-step
analyses.

Figure~\ref{fig:component-silent-errors} shows that the score
reductions are accompanied by more silent errors on every benchmark. Across all 1,054
tasks, removing evidence-preserving exploration increases the count from $289$ to $409$, while removing
runtime validation increases it to $377$.

\subsection{Efficiency and Backbone Robustness}
\label{sec:efficiency-backbone}
\label{sec:policy-model}

\noindent\textbf{Efficiency.} Table~\ref{tab:main} shows that under KramaBench Full Input with
DeepSeek-V4-Flash, \system costs $\$1.16$ and averages
$89$ seconds per task, versus $\$6.38$ and $636$ seconds for smolagents DR, the strongest baseline
by Overall score. Under Oracle Input, it costs $\$0.91$ and averages $67$ seconds, versus
$\$0.93$ and $245$ seconds for DS-STAR, the strongest baseline in this setting.

\noindent\textbf{Backbone robustness.} To assess performance across backbone models, we additionally
evaluate all systems with GPT-5.6 Luna and GPT-4.1-mini under KramaBench Full Input.
The inclusion of the earlier GPT-4.1-mini model is motivated by concerns about benchmark contamination.
As shown in Table~\ref{tab:main}, \system achieves Overall scores of $0.738$ and $0.460$,
respectively, outperforming all evaluated baselines with both backbones.

\section{Related Work}
\label{sec:related}

\paragraph{Data-science agents.}
Data-science agents aim to automate end-to-end analysis. Benchmarks cover scientific discovery,
multi-step reasoning, deep data research, and table question answering~\citep{scienceagentbench,dabstep,
dabench,dsbench,deepdataresearch,dsgym,tablebench}, while broader studies survey the area and examine
the limitations of open models~\citep{dataagentsurvey,opensourceanalysis}. Existing systems support
visualization~\citep{matplotagent}, insight discovery and report generation~\citep{agentada,datastorm,
insightpilot,dagent,toolaugmentedda}, heterogeneous data analysis~\citep{dsstar,datacross,agenticdata,
datacopilot}, and general data-science workflows~\citep{datainterp,datawiseagent,dsagent}. Recent work
trains open models for general data analysis or process verification~\citep{datamind,deepanalyze,
dataprm}, while EvoDS develops reusable skills and long-horizon context~\citep{evods}. Adapting
fine-tuned methods to new domains or model backbones may require additional data and retraining, and they may
not apply to proprietary API models, motivating \system's training-free design. Closest to our
setting, DS-STAR and Blackboard use file analyzers or partition agents to access heterogeneous
collections~\citep{dsstar,blackboardds}. Their downstream outputs may omit the source location, scope,
or coverage needed to audit later choices; \system preserves this context in evidence.

\paragraph{Runtimes for AI-driven analytics.}
Work on AI-driven analytics covers LLM-backed maps, filters, joins, and aggregations over unstructured
data, as well as execution optimization for cost and output quality~\citep{palimpzest,ipdb,lotus,abacus,docetl,quite}.
AOP predefines semantic operators and uses LLMs to compose and interactively execute pipelines for
complex data-lake queries~\citep{aop}. \citet{russo} connect optimized semantic-operator execution with
Deep Research agents through agent-backed search and compute operators and reusable materialized
contexts. Across these systems, explicit operators support pipeline construction and optimized
execution. In \system, typed operators make agent-selected data operations, their dependencies, and
intermediate states observable as execution records for runtime validation.

\section{Conclusion}
\label{sec:conclusion}

We present \system, an auditable runtime for agentic data analysis.
It preserves source locations, observation scope, and coverage, and links typed operations to runtime
observations so that conflicts can prompt diagnostic feedback and revision.
Across KramaBench, DA-Code, and DABStep, \system achieves the highest Overall scores among
the evaluated agents, with $28.6\%$ fewer observed silent-error cases than DS-STAR.
On KramaBench, it also costs less than the strongest baseline in each input regime.
Runtime validation remains limited by available evidence; broader validation coverage and
cost-based optimization are directions for future work.

\bibliographystyle{plainnat}
\bibliography{reference}

\end{document}